\documentclass[10pt, conference, compsocconf]{IEEEtran}
\usepackage{cite}
\usepackage{mathtools}
\usepackage{gensymb}
\usepackage{tabu}
\usepackage[bf]{caption}
\usepackage{amsmath}
\usepackage{algorithm}
\usepackage[noend]{algpseudocode}
\usepackage{breqn}
\usepackage{subcaption}
\usepackage{caption}
\usepackage{float}
\usepackage{float}

\begin{document}
\title{HICFR: Real Time 3D Indoor Human Image Capturing Based on FMCW Radar}
\author
{\IEEEauthorblockN{Hanqing Guo, Nan Zhang, Saeed AlQarni, Shaoen Wu} 
\IEEEauthorblockA{Ball State University\\
Muncie, IN USA \\
\{hguo, nzhang, saalqarni, swu\}@bsu.edu
}
}

\maketitle

\begin{abstract}
Most smart systems such as smart home and smart health response to human's locations and activities. However, traditional solutions are either require wearable sensors or lead to leaking privacy. This work proposes an ambient radar solution which is a real-time, privacy secure and dark surroundings resistant system. In this solution, we use a low power, Frequency-Modulated Continuous Wave (FMCW) radar array to capture the reflected signals and then construct to 3D image frames. This solution designs $1)$a data preprocessing mechanism to remove background static reflection, $2)$a signal processing mechanism to transfer received complex radar signals to a matrix contains spacial information, and $3)$ a Deep Learning scheme to filter broken frame which caused by the rough surface of human's body. This solution has been extensively evaluated in a research area and captures real-time human images that are recognizable for specific activities. Our results show that the indoor capturing is clear to be recognized frame by frame compares to camera recorded video.
\end{abstract}

\section{Introduction}\label{sec : intro}
Indoor human image capturing is of utmost important to many intelligent devices or systems. For example, robots need real-time human images to plan and change the route, and smart health system needs human images to recognize their activities thus alert when children or elderly people fall. However, most human image capturing solutions based on cameras, which makes users concern about privacy leaking problem\cite{townsend2011privacy,kumar2014survey}. Hence, human image capturing without computer vision technology has been a very popular research topic.

While traditional camera-based solutions result in privacy issues, the wearable sensors are devised to collect and process human motion data in many smart home scenarios. However, those wearable devices are inconvenient to users because they have to remember to equip these sensors when wakeup, and take off it when typing, wash hands or strenuous exercise\cite{stankovic2005wireless}. The worse thing is when they take off those wearable sensors, and the capture process would keep still; thus the results are unreliable at that time. Hence then, it is highly demanded to design a passive, non-invasive and real-time indoor human image capture solution so that privacy protection and convenience issue can be guaranteed. There are two more benefits to use radar or Radio Frequency (RF) technology to capture indoor human activities. One is RF solution can "see" human in dark light condition, while the other one is RF signals can sense human activities through the wall.

In this paper, we propose a scheme of \underline{H}uman \underline{I}mage \underline{C}apturing based on \underline{F}MCW \underline{R}adar sensors, HICFR. This work uses a FMCW radar to sense environment, then convert raw signals to 3D human images which contain spacial location information of target in real-time. It has the following highlights:

  $\bullet$ It uses antenna array combined with FMCW radar to send and receive directional beamforming to sense 3D environment, the frequency is 3.3GHz to 10GHz, and it is a very low power system, with average transmit power is below -40dbm/MHz. 
  
  $\bullet$ It has data clean feature, and the proposed calibration algorithm can record static environment response, then remove them from raw signals, thus only useful human motion responses can be reserved, which makes visualization results more clear.
  
  $\bullet$ It uses deep learning algorithm to process raw 3D images, and the deep learning model is trained to recognize whether the current frame is caused by irregular reflections, if true, filter them out so that the real-time captured frame is continuous.
  
  $\bullet$ Its design leverages feasible low-cost devices and achieves reliable performance in real-time applications.

In the rest of this paper, Section \ref{sec : relate} reviews the literature solutions related to our work. Next, Section \ref{sec : overview} describes the overview system structure, includes system platform introduction, signal processing chains as well as surrounding reflection signals removing. Then, Section \ref{sec : image} proposes the technical details to combine FMCW radar with antenna array to collect 3D array which represents the power at specific spacial voxels. Section \ref{sec : reflection} presents a novel solution that uses deep learning to make our scheme recognize and remove bad reflection frames, followed by Section \ref{sec : performance} evaluates the performance and feasibility of the whole system.

\section{Related work}\label{sec : relate}

Related works in this field include traditional sensors solutions, computer vision solutions and RF solutions. Recently, all of those solutions have been investigated for many smart home applications.
\subsection{Sensors and Computer Vision solutions}
Accelerometer and Gyroscope\cite{mukhopadhyay2015wearable} are the most common sensors applied to collect human motion data. Besides that, Inertial measurement unit (IMU) sensor which combines accelerometer and gyroscope sensor is also widely used to wearable devices\cite{ahmad2013reviews}. Those sensors can collect linear acceleration, rotation angle, angular velocity of targets, so that human who wears sensors can be collected motion data. Based on raw data collected by those sensors, researchers proposed various algorithms to recognize human activities\cite{zhang2011feature,zhang2012motion,attal2015physical,hammerla2016deep}. Zhang et al. designed physical features based on physical parameters of human motion, then find the most critical physical features for human activities, thus to improve the recognition accuracy\cite{zhang2011feature,zhang2012motion}. Years later, Ferhat et al. investigated k-Nearest Neighbor (k-NN), Support Vector Machines (SVM), Gaussian Mixture Models (GMM) classification techniques to process raw data and found the best scheme to recognize human activities\cite{attal2015physical}. Recently, researchers from the UK explored how to use deep, convolutional and recurrent models to detect human activities\cite{hammerla2016deep}.
In the meantime, many algorithms have been proposed based on computer vision techniques\cite{sung2011human,jalal2012depth,xia2011human,munaro2016rgb,chang2017feature}. Sung et al. used RGB-D images to detect and track human motions, those images with depth information were properly processed to achieve capture purpose\cite{sung2011human,munaro2016rgb}. Jalal's team proposed a solution, which uses translation and scaling invariant features with depth videos to recognize human logging activity\cite{jalal2012depth}. More recently, Kinect was widely used to collect human motion data because of its abundant APIs; researchers trained mechine models to do image segmentation for Kinect real-time video and capture coarse human outlines and motions\cite{xia2011human,chang2017feature}.

\subsection{RF Solutions}
Radar sensors and RF devices usually used for military or wireless communication purpose. However, it has been recently considered for smart home applications because of its data confidentiality, and the performance does not depend on lighting conditions. Recent researches either based on FMCW radar \cite{adib2013see, adib20143d, adib2015capturing, zhao2018through} or used off-the-shelf devices \cite{zhu2018distance, zhu2018indoor, avrahami2018below}. A research group of Massachusetts Institute of Technology (MIT) Adib et al.  designed MIMO antenna sensor with FMCW technique to detect human move \cite{adib2013see} and even capture human figure through a wall \cite{adib2015capturing, adib20143d}. Off-the-shelf devices such as ultrasonic sensor or walabot \cite{walabot} were also investigated feasibility to recognize human activities \cite{zhu2018distance,zhu2018indoor,avrahami2018below}. Avrahami et al. proposed a human activity recognition scheme based on 2D heat maps generated by walabot, while Zhu et al. \cite{zhu2018indoor} applied traditional signal processing algorithms to filter and cluster raw data thus recognize human actions. Both of them achieve higher than 80\% accuracy in their research tasks.

\section{System Design Overview}\label{sec : overview}
To enable real-time human image capturing with ambient radar in smart home scenario, we propose a solution: \underline{H}uman \underline{I}mage \underline{C}apturing based on \underline{F}MCW \underline{R}adar sensors (\textit{HICFR}) system. \textit{HICFR} scans 3D surroundings reflections with FMCW chirps and 2D antenna array. While FMCW chirps are used to compute direct distance from the detected object to receive antenna, and 2D antenna array is placed to identify spatial directions. It emits FMCW chirps to scan 3D volume of surroundings, then received signals are processed to remove environment fix reflections,. After that, we calculate the reflection powers of any scanned voxel and construct it to 3D images, then a Deep Neural Network (DNN) based filter algorithm is designed to address noncontinuous reflection frames problem, thus capture real-time human activities.

\subsection{Sensing Platform}
\begin{figure}[h]
\centering
\includegraphics[width=.4\textwidth]{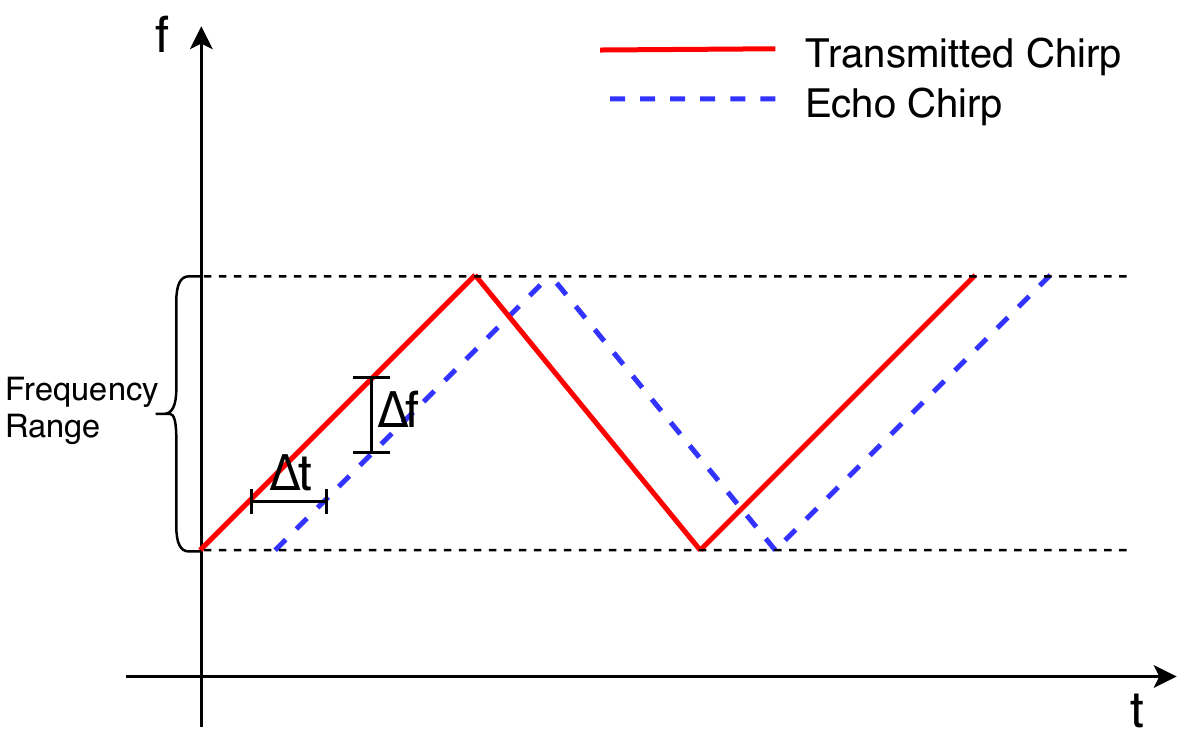}
\vspace{-0.1in}
\caption{Frequency Chirp}
\vspace{-0.1in}
\label{fig: fmcw}
\end{figure}

Since \textit{HICFR} requires to emit FMCW chirps and collect received signals by 2D antenna array, there is an off-the-shelf radar sensor called Walabot \cite{walabot} meets our requirement. Walabot has compact size and low-cost feature with a board size of $72mm*140mm$ and the average power is lower than -41dbm/MHz. The frequency range of FMCW chirp emitted by Walabot is 3.3GHz-10GHz, which is good enough to detect direct distance within 10 meters range based on gradient of FMCW chirp. It also contains 18 pair of antenna, which are arranged to 2D antenna array. 
\begin{figure}[h]
\centering
\includegraphics[width=.5\textwidth]{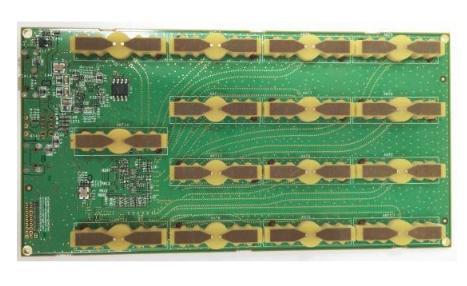}
\vspace{-0.3in}
\caption{2D Antenna Array of Walabot}
\vspace{0.0in}
\label{fig: real_antenna}
\end{figure}
Figure \ref{fig: real_antenna} shows internal antenna array of Walabot. Walabot emits FMCW chirps to scan $\phi$ in horizontal direction and $\theta$ in vertical direction. Then it communicates to \textit{HICFR} with USB port to send raw signals for further processing. The scanned area of walabot can be present as figure\ref{fig: axis} below: 
\begin{figure}[h]
\centering
\includegraphics[width=.5\textwidth]{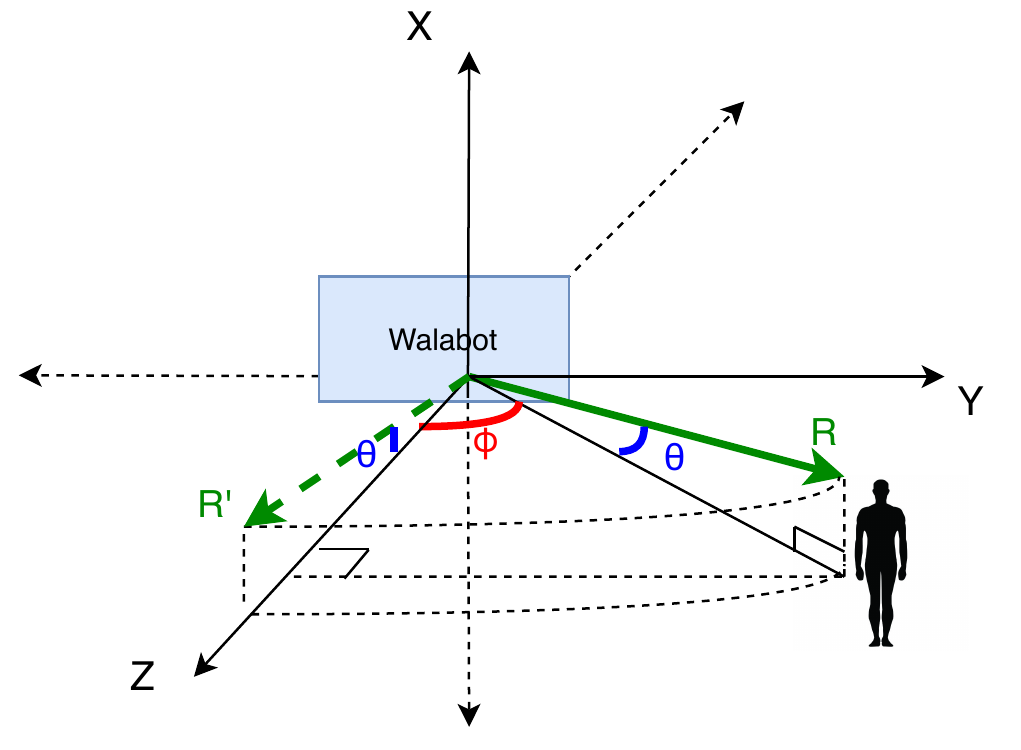}
\vspace{-0.3in}
\caption{Scanned 3D Axis of Walabot}
\vspace{0.0in}
\label{fig: axis}
\end{figure}
Where $\theta$ is Elevation angle to detect the height of human, and $\phi$ is Wide angle to capture the width of human. $R$ is FMCW signals travel distance from transmit antenna to humans head, and $R'$ is hypotenuse of triangle whose angle is $\theta$ and hypotenuse $R$ rotate $\phi$ degree, the scan range is the sector which triangle passed. In our case, $\theta$ is from $-45\degree$ to $45\degree$ and $\phi$ is from $-90\degree$ to $90\degree$. The direct travel distance $R$ can be calculated by FMCW properly with formula (\ref{equ: fmcw_r}) and figure \ref{fig: fmcw} as below:
\begin{equation}
R = \frac{c|\bigtriangleup t|}{2}=\frac{c|\bigtriangleup f|}{2(df/dt)}
\label{equ: fmcw_r}
\end{equation}

While $\bigtriangleup t$ is signal travel time from transmit antenna to object and reflect back to receive antenna, and $\bigtriangleup f$ is frequency difference of transmit and receive signals. $df/dt$ is slope of transmit or echo frequency chirp and $c$ is speed of light. To simplify description, Equation (\ref{equ: fmcw_r}) and formula \ref{fig: fmcw} are not considering doppler frequency shift effect.

\subsection{System Flowgraph}
\begin{figure}[h]
\centering
\includegraphics[width=.4\textwidth]{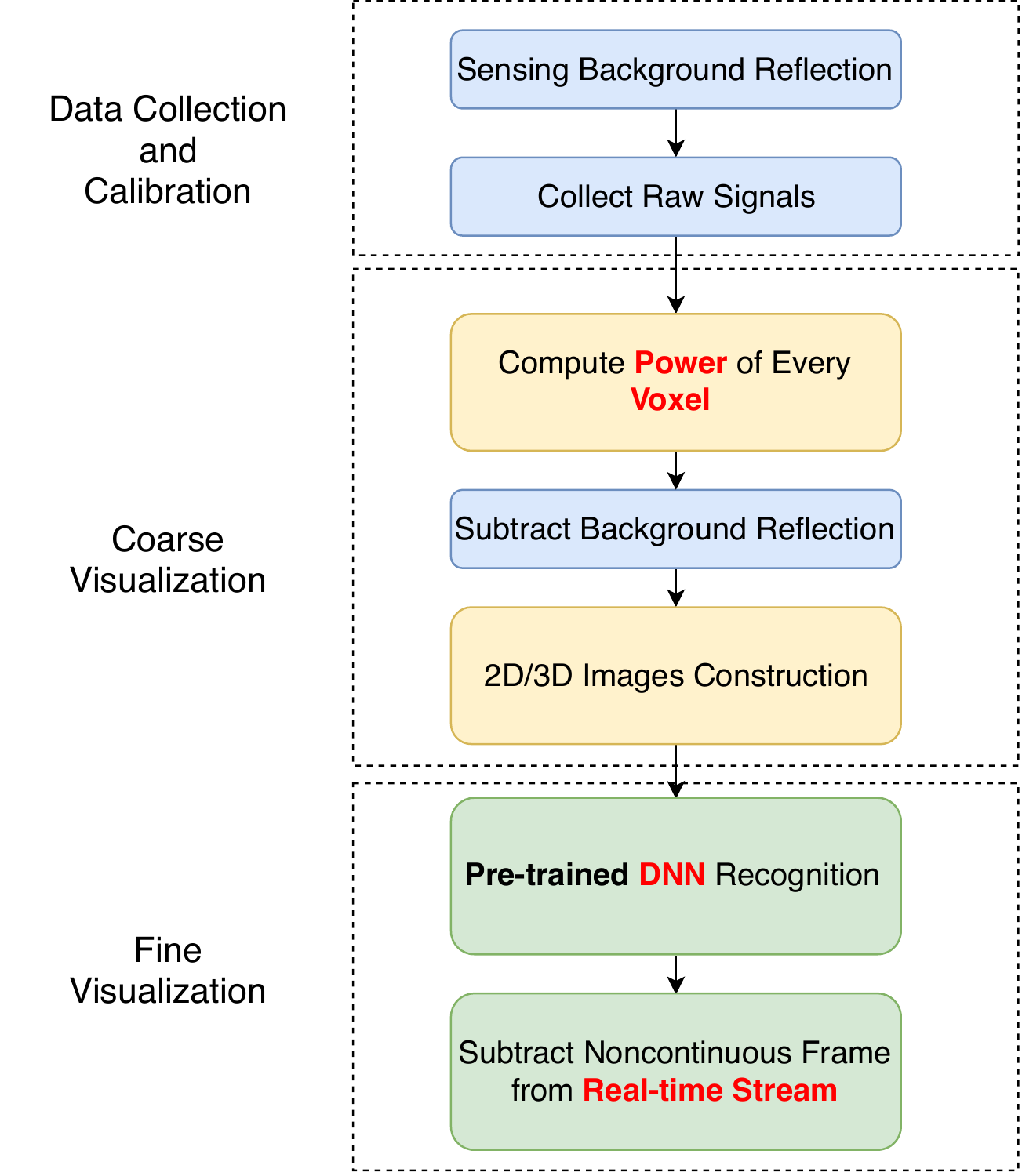}
\vspace{-0.0in}
\caption{Flow Graph of \textit{HICFR}}
\vspace{-0.1in}
\label{fig: flow}
\end{figure}

The complete \textit{HICFR} system contains 3 phases. 1) Data collection and Calibration, 2) Coarse Visualization and 3) Fine Visualization. As shown in figure \ref{fig: flow}, environment sensing period is running ahead of detecting. First, \textit{HICFR} emits FMCW chirps and records static background reflections, when \textit{HICFR} starts to do object detection task, its 2D antenna array collects raw signals. Second phase is designed to convert signals to images. Since \textit{HICFR} scans 3D surroundings with parameters $R, \theta$ and $\phi$, the raw received signals then being processed to represent power of every spatial points with different $R, \theta$ and $\phi$ , namely voxel in scanned area. Then \textit{HICFR} subtracts recorded background reflections power, and removes environment fix reflection to get pure power information of changed objects, which is a 3D matrix $M$ with dimension of $(sizeX,sizeY,sizeZ)$, where $sizeX, sizeY, sizeZ$ can be computed by equations (\ref{equ: dim}).
\begin{equation}
\vspace{-0.0in}
\begin{split}
&sizeX = range(R)/ res(R) \\
&sizeY = range(\theta)/ res(\theta) \\
&sizeZ = range(\phi)/ res(\phi)
\end{split}
\label{equ: dim}
\end{equation}

In the equations above, $range(\cdot)$ is detect range of parameters, while $res(\cdot)$ is designated parameters' sampling  interval. Next, we propose a novel solution to achieve coarse-to-fine visualization. Because human body acts as an uneven reflector rather than a scatterer, thus some signals reflect back directly to antenna array, while other signals are deflected from normal path or even away from receive antenna, in this case, constructed 3D images may contain some ambiguous results. The third phase addresses this issue by using Machine Learning algorithm. We collect dataset for regular reflection and deflected reflection images, then train a Deep Neural Network (DNN) which contains Convolutional Layer, Pooling layer and Linear layer to recognize them. The trained DNN is placed to main program loop thus eliminates ambiguous frames from real-time stream.

As shown above, the key challenges and main contributions of \textit{HICFR} are 1) Compute reflection powers of every voxel with $R, \theta$ and $\phi$ based on the received complex signals of antenna array, 2) Construct 2D/3D images with known 3D power matrix, 3) Address ambiguous images issue which caused by signal deflection with Deep Neural Network and achieve real-time filtering scheme.

\section{Calibration and Visualization}\label{sec : image}
In this section, we dive in the technical detail of \textit{HICFR}. Since walabot antenna array collects RF-signals, which is complex signals, they can be represented by amplitude and phase as follows:
\begin{equation}
s_t = A_te^{-j2\pi\frac{r}{\lambda}t}
\label{equ: original_signal}
\end{equation}
Where $s_t$ is signals received at $t$ moment. $A_t$ is amplitude of signal at time $t$, $r$ is travel distance of signal and $\lambda$ is signals' wavelength. Since received phase has linear function with travel distance, so $2\pi\frac{r}{\lambda}t$ is the signal phase when it reach to receive antenna at moment $t$.

Revisit to equation \ref{equ: original_signal}, due to the receiver is an antenna array, so $s_t$ should have more complex format to specify the signal is received by which receive antenna, note as $s_{n,t}$, where $n$ is $n_{th}$ antenna number, and $s_{n,t}$ is signals received by antenna $n$ at moment $t$. 

\begin{figure}
    \centering
    \begin{subfigure}[b]{0.23\textwidth}
        \includegraphics[width=\textwidth]{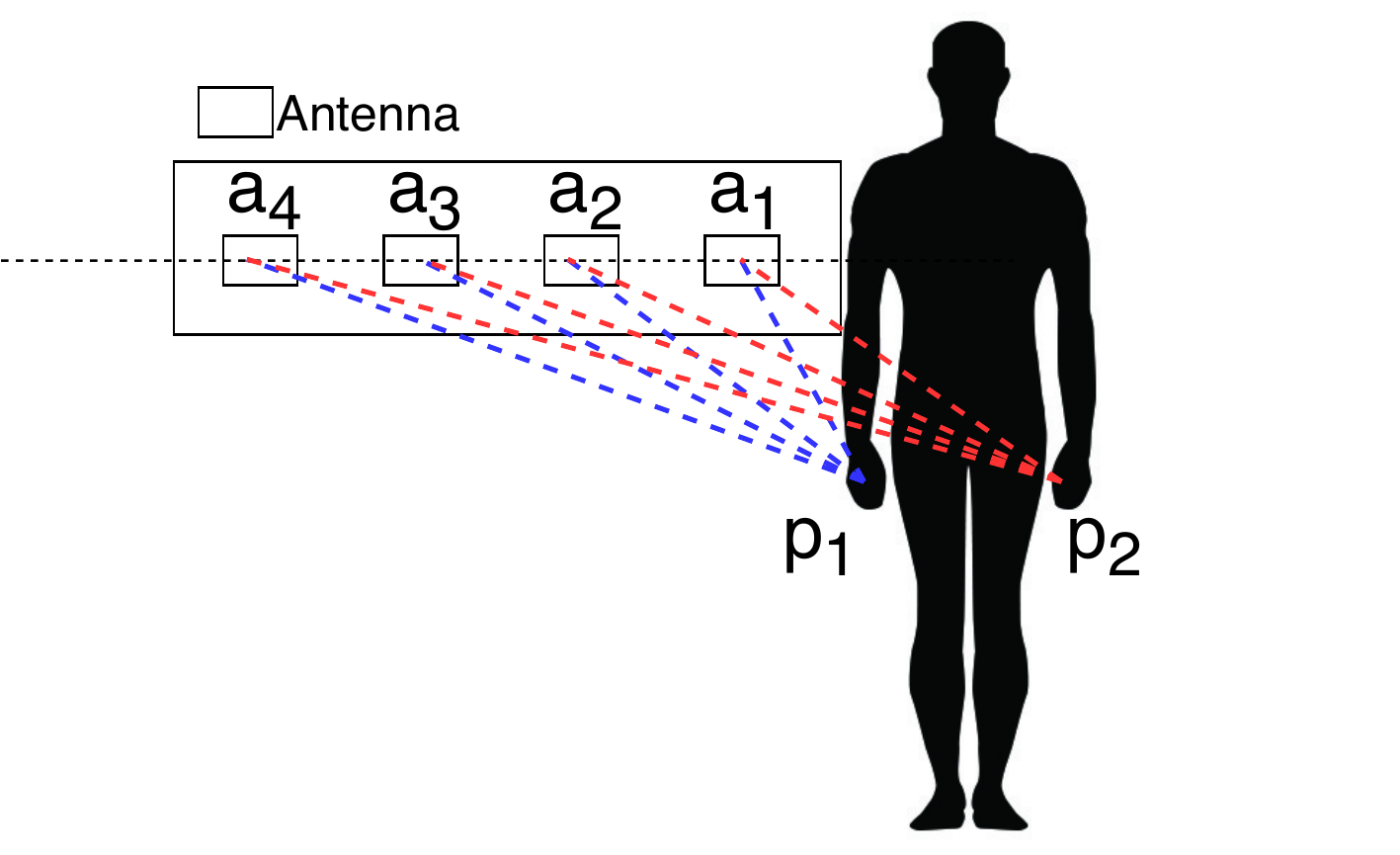}
        \caption{Multiple Points}
        \label{fig: multi-points}
    \end{subfigure}
    \hspace{-0.5in}
    ~ %add desired spacing between images, e. g. ~, \quad, \qquad, \hfill etc. 
    \begin{subfigure}[b]{0.3\textwidth}
        \includegraphics[width=\textwidth]{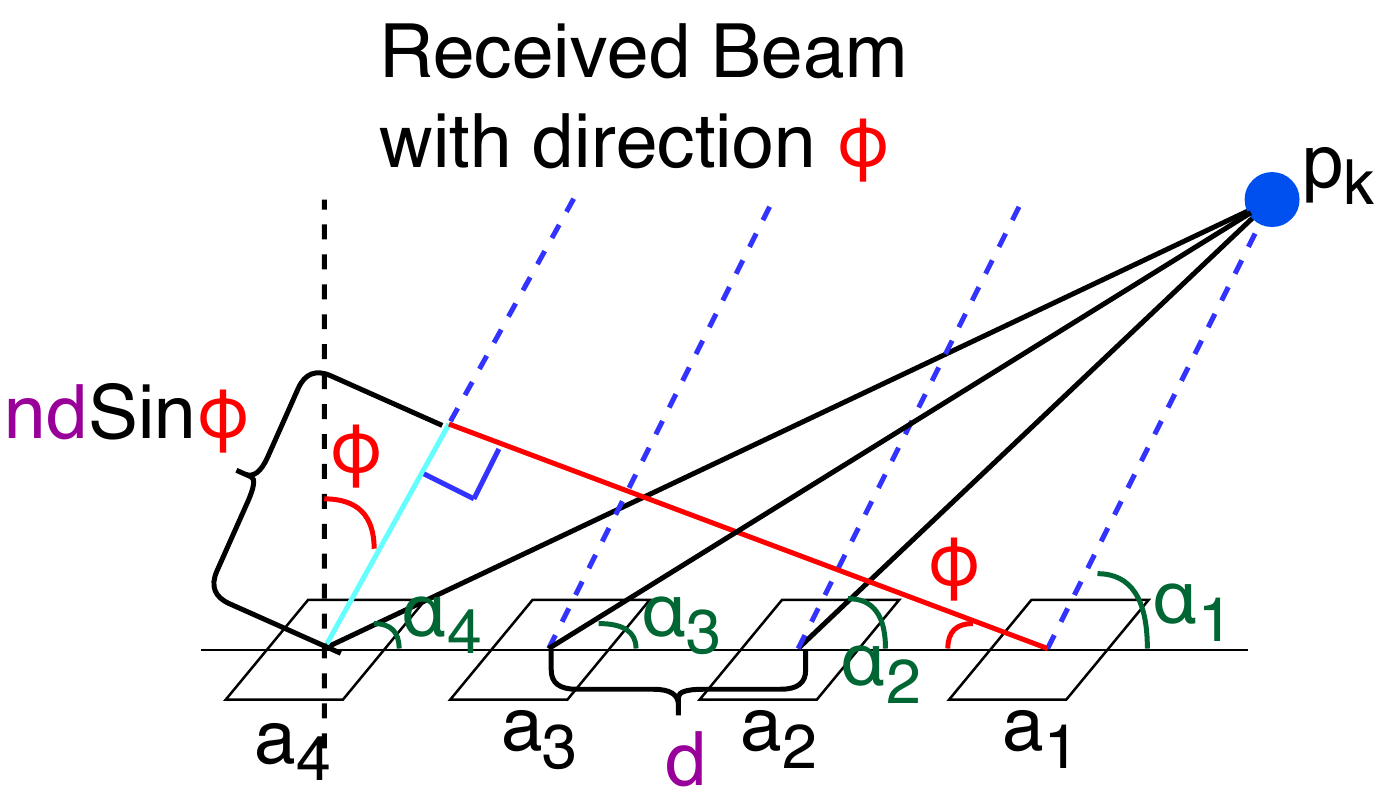}
        \caption{2D shape}
        \label{fig: 2D shape}
    \end{subfigure}
    \caption{2D Scanning Scenario}
    \vspace{-0.25in}
\end{figure}

Another parameter needs to be clarified is $r$. Since human body is a surface rather than a point, it reflects signals from different directions to all antennas, the received signals at moment $t$ of one antenna contains more than one points' reflection, thus $r$ varies from multiple reflect points. Figure \ref{fig: multi-points} shows when antenna array scans human body, his left hand $p_1$ reflects to antenna $a_1, a_2, a_3, a_4$ as blue dot line, his right hand $p_2$ reflects to antenna array as red dot line. Based on above description, equation \ref{equ: multi-antenna} is designated as follow:
\begin{equation}
\begin{split}
&s_{n,t} = \displaystyle\sum_{k=1}^K A_{n,t}e^{-j2\pi\frac{r_{n,k}}{\lambda}t}\\
&r_{n,k} = travel (p_k, a_n)
\end{split}
\label{equ: multi-antenna}
\end{equation}
Suppose $p_k$ is $k_{th}$ points on the detected object, then $K$ is number of points being scanned, $r_{n,k}$ is signal travel from $p_k$ to antenna $a_n$. 

\subsection{Compute Voxel Power}
\hspace{-0.15in} \textbf{Power of Direction:} Based on equation \ref{equ: original_signal} and \ref{equ: multi-antenna}, the problem can be decleared as: known signals $s_{n,t}$ received by antenna $a_n$ at moment $t$, then compute reflection power of every scanned points. Because both angles and distance property can be reflected to phase of received signal. More specifically, the power of specific angle $\phi, \theta$ can be refered by antenna array property, while the power of specific distance $r$ can be calculated by FMCW feature. Revisit to figure \ref{fig: multi-points} and change antenna array panel to a plane figure, antenna $a_1, a_2, a_3, a_4$ receive reflection from $p_k$, the coming direction of beam is $\phi$ as shown in both figure \ref{fig: 2D shape} and figure \ref{fig: axis}. While $\alpha_1, \alpha_2, \alpha_3, \alpha_4$ are angles between antenna to $p_k$, and $d$ is distance between two antennas. Thus power of direction $\phi$ can be presented as $P(\phi)$ in equation \ref{equ: angle}:
\begin{equation}
P(\phi) = |\displaystyle\sum_{n=1}^N s_{n,t}e^{-j2\pi\frac{nd\sin\phi}{\lambda}}|
\label{equ: angle}
\end{equation}
Where $N$ is how many antennas in the dimension. Because $s_{n,t}$ travel different distance for each antenna, and the difference can be represented by $nd\sin\phi$ as depicting with light blue color. Thus their phase change of antenna $n$ is $2\pi\frac{nd\sin\phi}{\lambda}$, $\lambda$ is signal wavelength.

\hspace{-0.15in} \textbf{Power of Distance:} The travel distance of signals also related to the direct distance from point $p_k$ to antenna $a_n$. Frequency Modulated Continuous Wave measures reflection depth by calculating frequecy shift between transmit and receive chirp. Equation \ref{equ: fmcw_r} shows the FMCW feature. We define $v$ is slope of frequency chirp versus time, where $v $ is equal to $df/dt$ in figure \ref{fig: fmcw}. So the power of distance $r_k$ can be calculated by phase change of $s_{t,n}$ as shown below in equation \ref{equ: pr}: 
\begin{equation}
P(r_k) = |\displaystyle\sum_{n=1}^N\sum_{t=1}^T s_{n,t}e^{-j2\pi\frac{vr_{n,k}}{c}t}|
\label{equ: pr}
\end{equation}
where $r_k$ is signal travel distance from point $k$. $T$ is the duration of each chirp. Because $f=vt$ and $r/c=t_{travel}$, we can easily get the phase change is $2\pi ft_{travel}$, thus power of $r_k$ is summation over duration $T$ and total antenna number $N$.

\hspace{-0.15in} \textbf{Power of Voxel:} Since \textit{HICFR} scans 3D surroundings, reconsider situation shown at figure \ref{fig: 2D shape}, where $p_k$ on same panel of antenna. However, points in 3D volume need three parameters to locate, either with ($r, \theta, \phi$) in spherical coordinate system or ($x, y, z$) in cartesian coordinate system. We choose spherical coordinate system because the power of $\theta$ and $\phi$ can be calculated based on our 2D antenna array. Figure \ref{fig: 3d} shows how it works:

\begin{figure}[h]
\centering
\includegraphics[width=.5\textwidth]{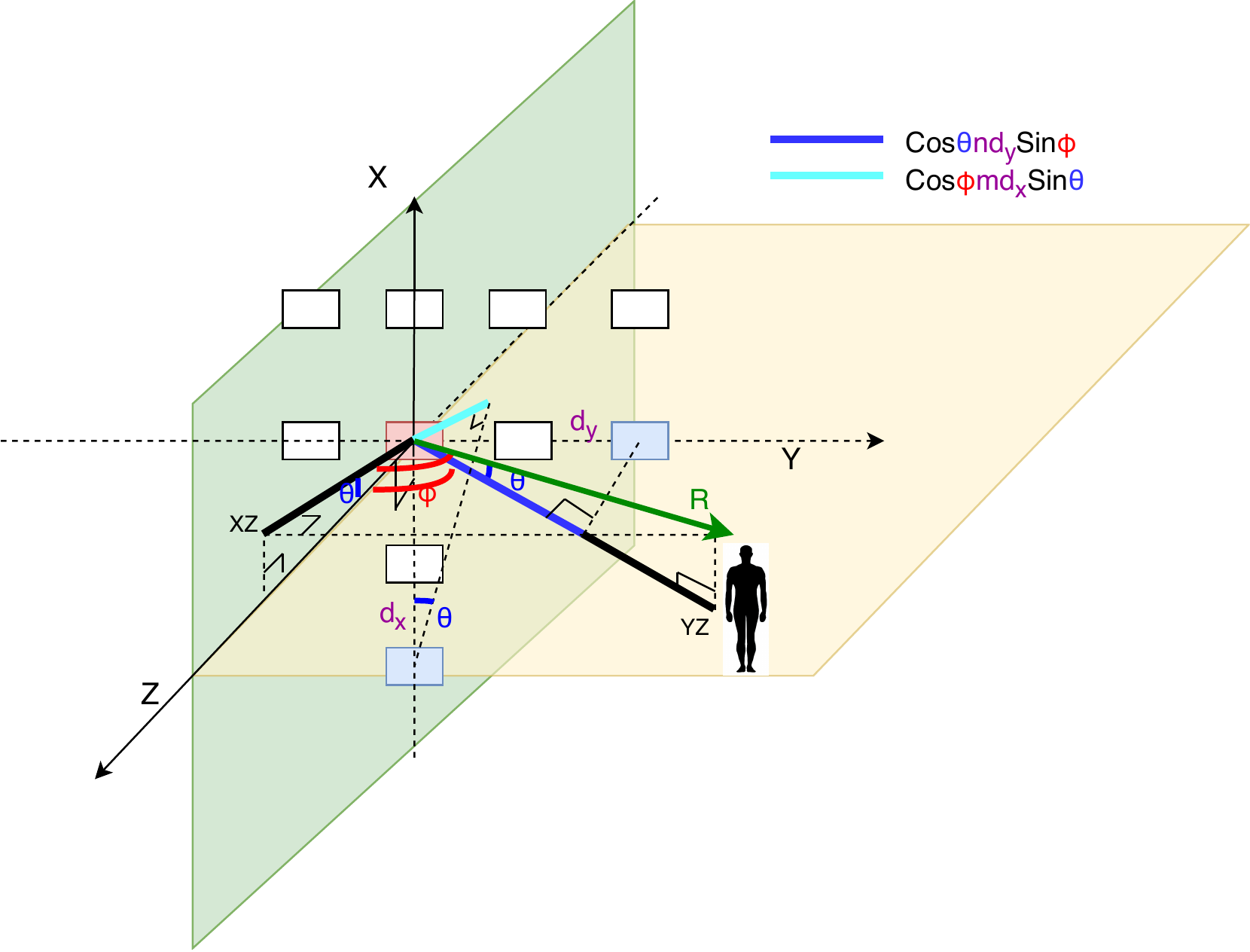}
\vspace{-0.15in}
\caption{3D Voxel Power Description}
\vspace{-0.1in}
\label{fig: 3d}
\vspace{-0.1in}
\end{figure}

The 2D antenna array is on $X-Y$ panel, where blocks is antenna. $d_x$ and $d_y$ are distance between two antenna in two dimensions. $Y-Z$ panel is the dimension drawn in figure \ref{fig: 2D shape}, and $d_y, \phi$ is $d, \phi$ in equation \ref{equ: angle}, while $\theta$ is elevation angle from $Y-Z$ panel to $R$. In 3D figure, $R$ is mapping to $Y-Z$ pannel as $YZ$ with $\cos\theta$, and it is mapping to $X-Z$ panel as $XZ$ with $\cos\phi$, where $\phi$ is wide angle from $YZ$ to $Z$ axis. Thus the distance change at $Y-Z$ panel for each antenna is $\cos\theta*nd_y*\sin\phi$ as blue line, that change at $X-Z$ panel is $\cos\phi*md_x*\sin\theta$ as light blue line shows.

\begin{equation}
\begin{split}
&P(r_k, \theta, \phi) = \\
&|\displaystyle\sum_{m=1}^M\sum_{n=1}^N\sum_{t=1}^T s_{n,m,t}e^{-j2\pi\frac{vr_k}{c}}e^{j\frac{2\pi}{\lambda}\cos\theta(nd_y\sin\phi+md_x\cos\phi)}|
\end{split}
\label{equ: total}
\end{equation}

Since distance change represents phase change of signals, then we can calculate power of any voxel by equation \ref{equ: total}. $s_{n,m,t}$ is the signal received by receive antenna $n$ from transmit antenna $m$ at time $t$.
\subsection{Construct 3D Image}
\hspace{-0.15in} \textbf{Remove Background Reflection:}
To get rid of environment reflections such as desks or walls, \textit{HICFR} starts a sensing process before capture humans, name as calibration. Since background reflection is static and the reflection power is fixed, so that calibration sensing, calculating and recording the background reflection power of any voxel, after that, when \textit{HICFR} starts human image capturing task, it subtracts the static background reflection power from the real-time reflection power. We need to make sure there is no human enters the lab during calibration period.

\begin{figure}
    \begin{subfigure}[b]{0.23\textwidth}
        \includegraphics[width=\textwidth]{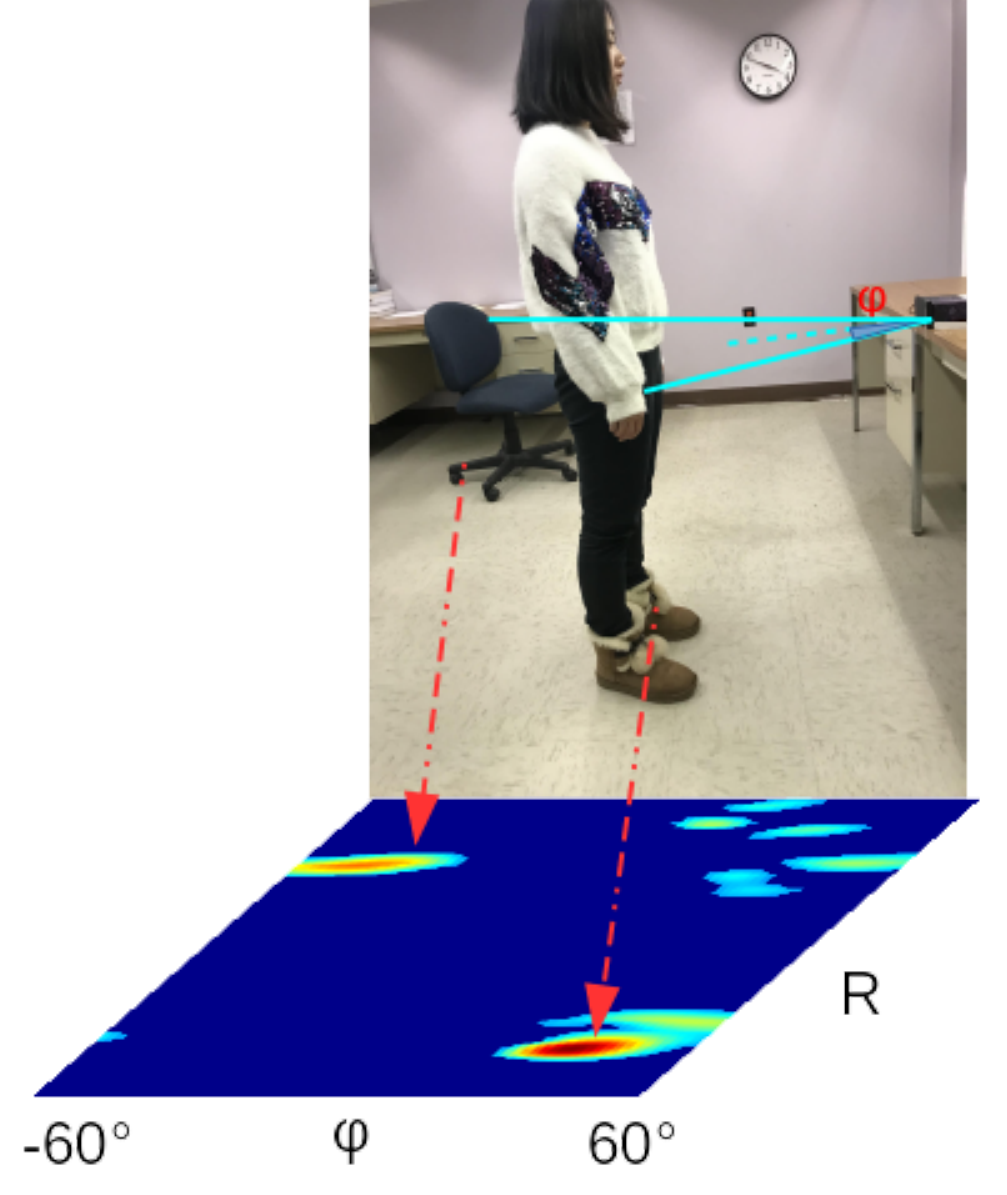}
        \vspace{0.2in}
        \caption{2D Heatmap}
        \label{fig: 2d-human}
    \end{subfigure}
    \hspace{-0.1in}
    ~ %add desired spacing between images, e. g. ~, \quad, \qquad, \hfill etc. 
    \begin{subfigure}[b]{0.22\textwidth}
        \includegraphics[width=\textwidth]{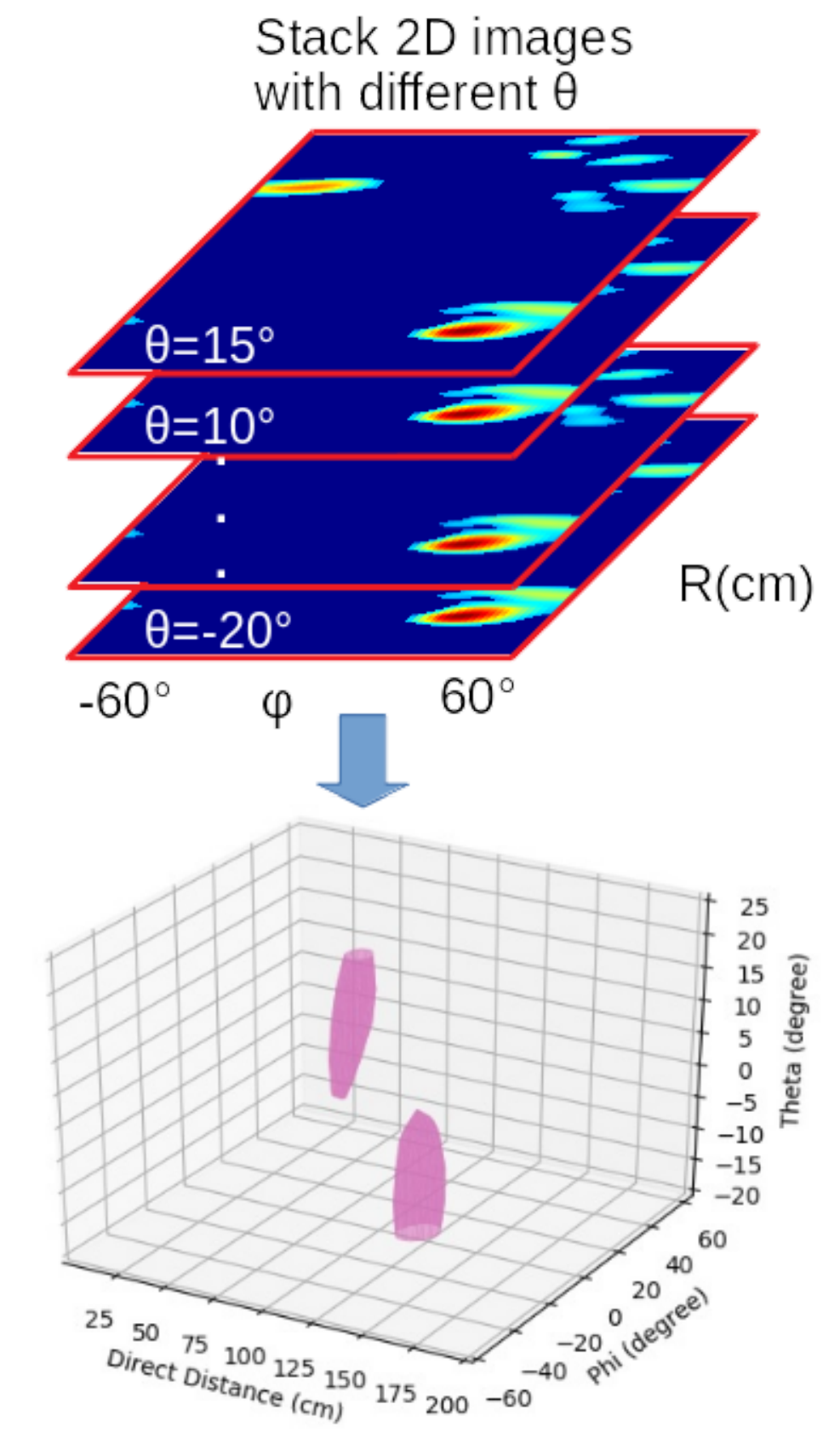}
        \caption{Stack to 3D Image}
        \label{fig: 3d-human}
    \end{subfigure}
    \caption{2D/3D Image Capturing}
\end{figure}

\hspace{-0.15in} \textbf{Construct 2D/3D Image:}
Once \textit{HICFR} calculates the power of every voxel and removes background reflection power, it gets a 3D matrix $M$ with the dimension of ($sizeX, sizeY, sizeZ$), where $sizeX, sizeY, sizeZ$ can be refered from equation \ref{equ: dim}.
Since a 2D image is related to either $(R, \theta)$, $(R, \phi)$ or even $(\theta, \phi)$. To make 2D image has a clear meaning, we choose to construct 2D image with distance and wide angle $(R, \phi)$. At first, we find the highest power from $M$, suppose the highest reflection power is from point at ($R_a, \theta_b, \phi_c$), then $M(R_a, \theta_b, \phi_c)$ is the highest value in $M$, and $M(R, \theta_b, \phi)$ is a 2D array because parameter $\theta$ is fixed as $\theta_b$. Thus we draw a 2D heatmap image based on $M(R, \theta_b, \phi)$, where the color shows reflection power intensity, the darker color means the higher reflection power at $(R, \theta_b, \phi)$. Figure \ref{fig: 2d-human} shows 2D image capturing scenario and its corresponding heatmap. 
\begin{figure}
	\vspace{-0.2in}
    \begin{subfigure}[b]{0.25\textwidth}
    \hspace{-0.1in}
        \includegraphics[width=\textwidth]{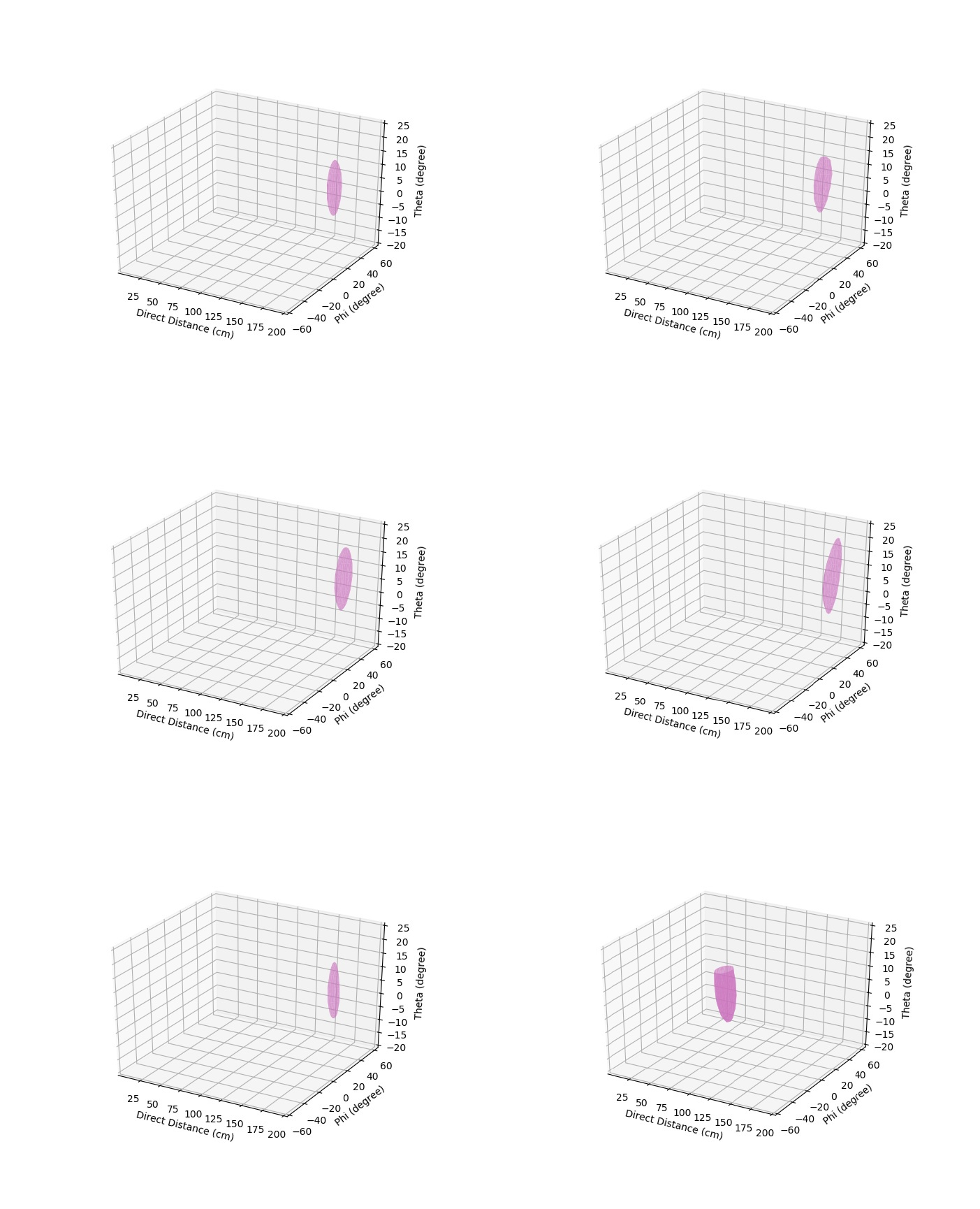}

        \caption{Regular Dataset}
        \label{fig: reg}
    \end{subfigure}
    \hspace{-0.4in}
    ~ %add desired spacing between images, e. g. ~, \quad, \qquad, \hfill etc. 
    \begin{subfigure}[b]{0.25\textwidth}
        \includegraphics[width=\textwidth]{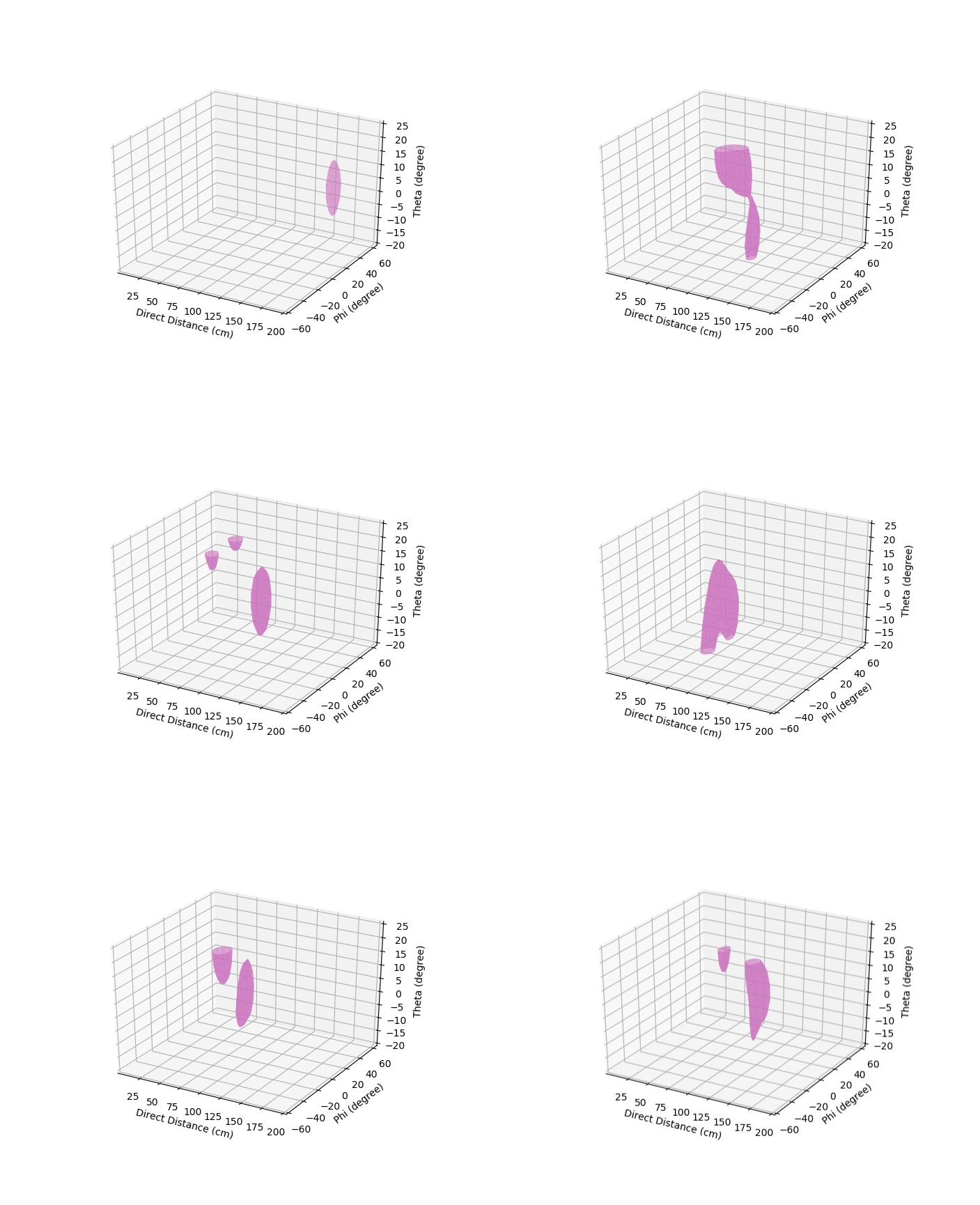}
        \caption{Ambiguours Dataset}
        \label{fig: amb}
    \end{subfigure}
    \caption{Dataset Overview}
    \vspace{-0.15in}
    \label{fig: dataset}
\end{figure}
As can be seen from figure \ref{fig: 2d-human}, the range of $\phi$ is from $-60\degree$ to $60\degree$, where $\phi$ is the angle from dash blue line to human, in this case, dash blue line is the base line in the middle of Walabot, thus $\phi$ is wide angle from  base line to object. While 2D image only depicts the highest power layer of fixed $\theta$,  3D image shows more information about object width, height and location. Figure \ref{fig: 3d-human} shows how to construct 2D images to 3D images. \textit{HICFR} uses marching cubes algorithm to draw vertices and faces of stacked images, then it uses nomarlized filter to remove low power points. It is very clear to see that human is at a shorter direct distance to radar, and the height of human is greater than the chairs in 3D vision.
\section{Filter Reflection}\label{sec : reflection}
Another challenge of real-time human image capturing is signal deviation. Since human body is not a plane surface, especially when human moves, the surface of body is extremely deformed. As a result, while our antenna array transmits siganls and scans human body, only signals that close to normal surface are reflected back toward the antennas. Other signals may be deviated from another routes and back to receiver, which makes our antenna "misunderstand" the real distance and angle from object. This scenario is shown in figure \ref{fig: misunderstand}:
\vspace{-0.05in}
\begin{figure}[h]
\centering
\includegraphics[width=.35\textwidth]{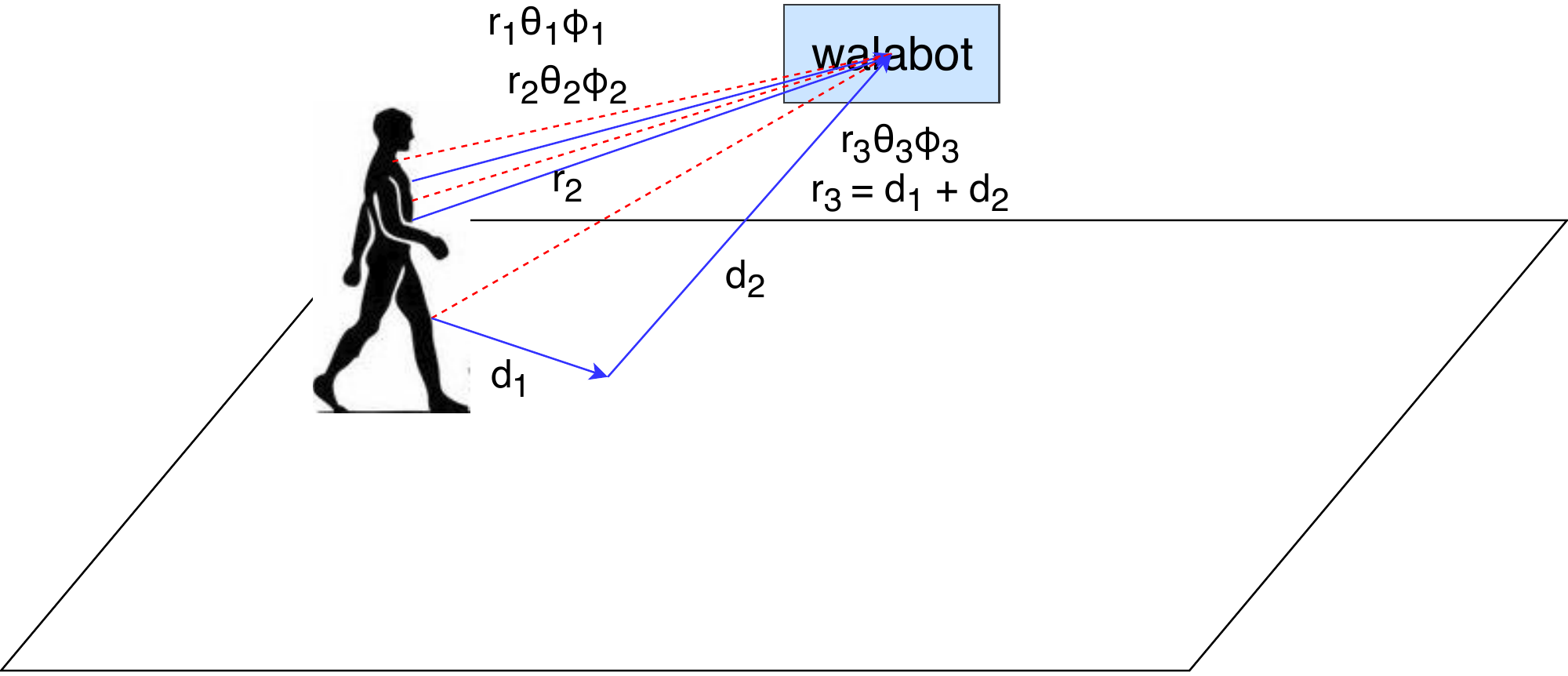}
\vspace{-0.05in}
\caption{Signal Deviation}
\vspace{-0.05in}
\label{fig: misunderstand}
\vspace{-0.1in}
\end{figure}

In this case, the distance between human chest and leg is not quite large, however, signals transmitted from antenna array travel to human leg and deviate it's coming route, thus receive antenna gets signals from $r_3\theta_3\phi_3$, where $r_3 = d_1 + d_2$, so that antenna "misunderstand" human leg position with wrong distance and angles, and it results in deformed 3D shape. To address this issue, we design a Deep Neural Network to recognize whether current 3D figure is deformed or not, and remove them from image capture stream.
\begin{figure*}
%\vspace{-0.1in}
\centering
\includegraphics[width=\textwidth]{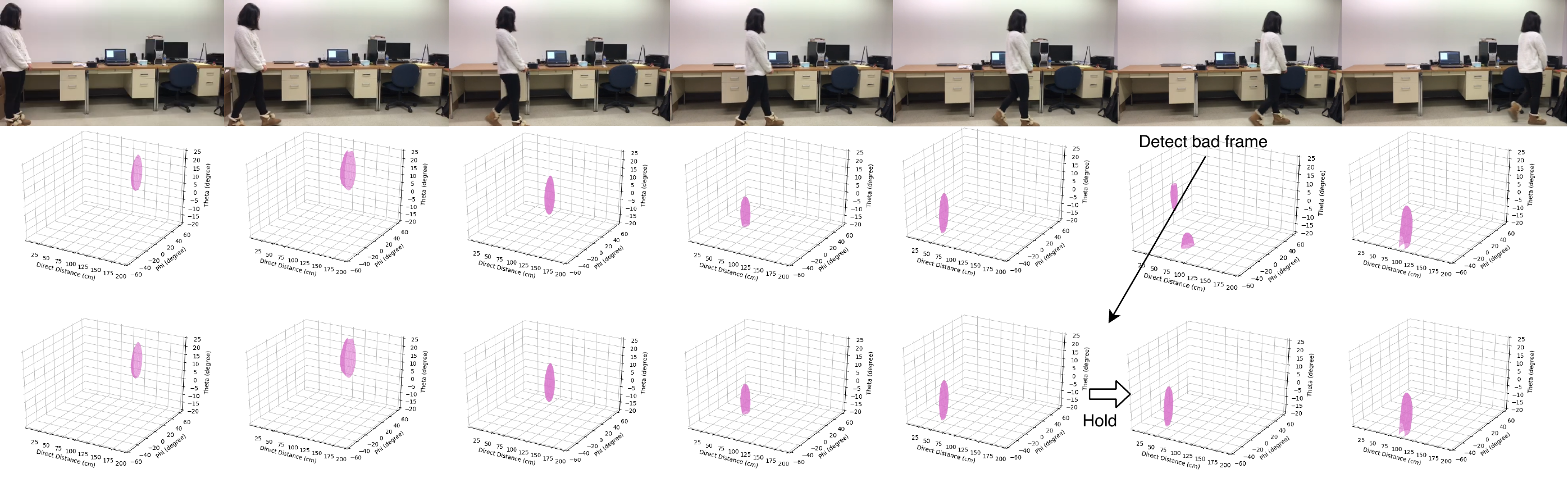}
\vspace{-0.1in}
\caption{Human Real-time Image Capturing} 
\vspace{-0.2in}
\label{fig: final}
\end{figure*}

\hspace{-0.15in} \textbf{DNN Recognization:} We use transfer learning technique to solve this problem more efficiently. Due to it's a image processing problem, the proper Deep Neural Network should have Convolutional layer to reduce possible parameters and amount of calculation. Based on that, we choose \textbf{resNet18} to classify our 3D image. Our contribution is 1) Collect regular and ambiguous images used as training dataset, 2) Change the network structure of resNet18 to make the DNN convergence faster, 3) Real-time load trained DNN parameters and handle recognization task in mainstream.

We collect training dataset from real human activities, while one person walks around in the lab, we construct 3D images and concatenate them as 3D videos. Then we classify them manually into 2 categories: regular frames and ambiguous frames. Figure \ref{fig: dataset} shows samples of dataset, while figure \ref{fig: reg} shows regular 3D reflection power and \ref{fig: amb} has ambiguous images. As can be seen from the dataset, the regular frames show  human 3D position very clear, and the ambiguous frames always "misunderstand" location of some part of human body. 

\hspace{-0.15in} \textbf{Change resNet18 Structure:} The first step to apply transfer learning is changing the last Fully Connect(FC) layer, the last FC layer dimension of normal resNet18 is $(512, 1000)$, which means $in\ feature$ to FC layer is $512$ and output $1000$ features. The $1000$ $out\ feature$ usually feed into $softmax$ functions to be classified into $1000$ categories. In our design, we only have $2$ categories: regular and ambiguous. Then we change the dimension of last FC layer to be $(512, 2)$. The second change of original resNet18 is changing the pooling layer before FC layer. Resnet18 uses Average Pooling layer to compress features to $512$, but Average Pooling sometimes cannot extract good features because it takes all into count and results an average value. Since our dataset images have strong edges, and \textbf{Max Pooling} extracts the most important or extreme features. So we change the pooling layer to the same size of Max Pooling layer and compare the different convergence of them.

\begin{figure}
    \begin{subfigure}[b]{0.25\textwidth}
    \hspace{-0.1in}
        \includegraphics[width=\textwidth]{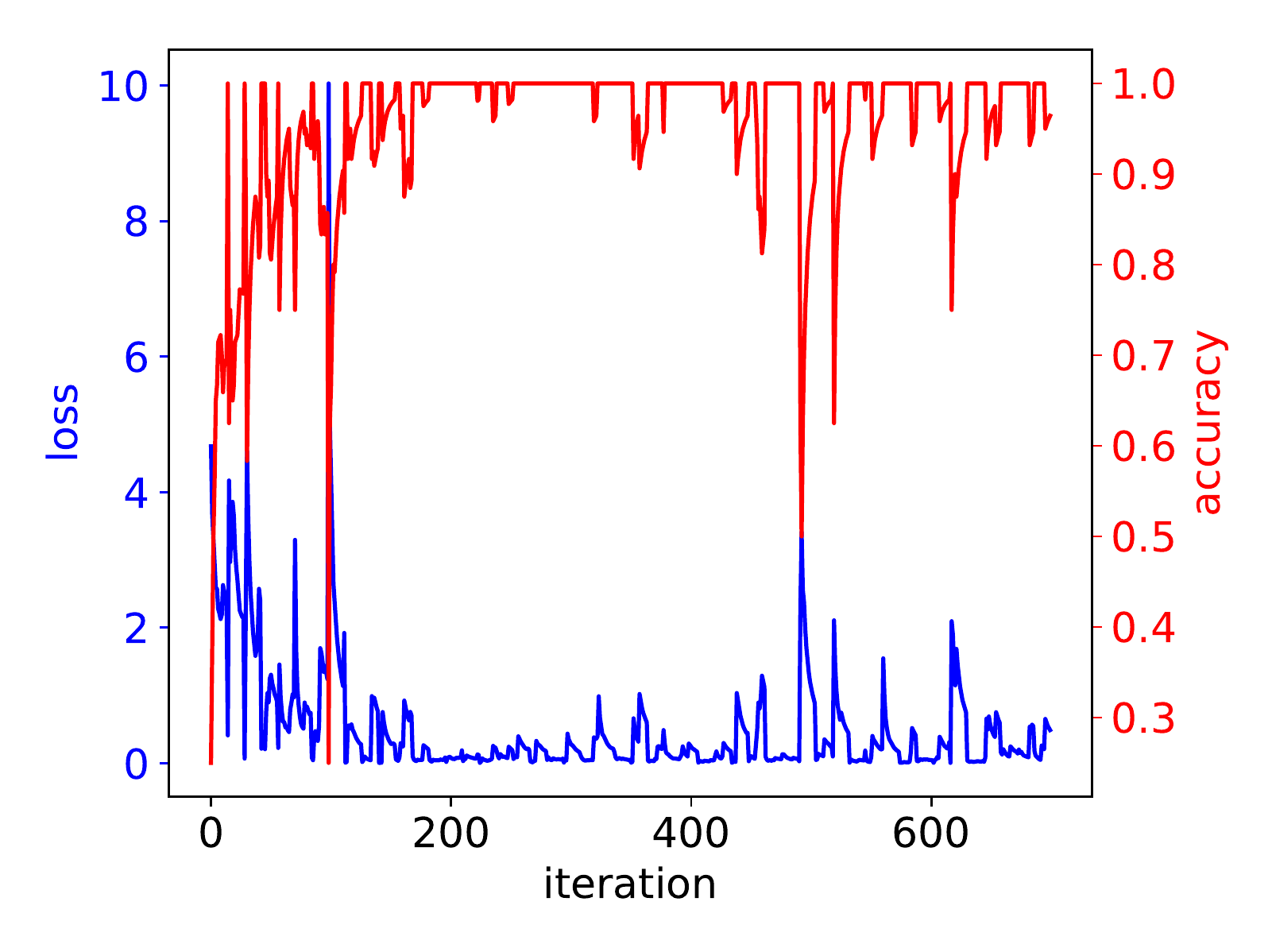}

        \caption{Average Pooling Result}
        \label{fig: avgpool}
    \end{subfigure}
    \hspace{-0.5in}
    ~ %add desired spacing between images, e. g. ~, \quad, \qquad, \hfill etc. 
    \begin{subfigure}[b]{0.25\textwidth}
        \includegraphics[width=\textwidth]{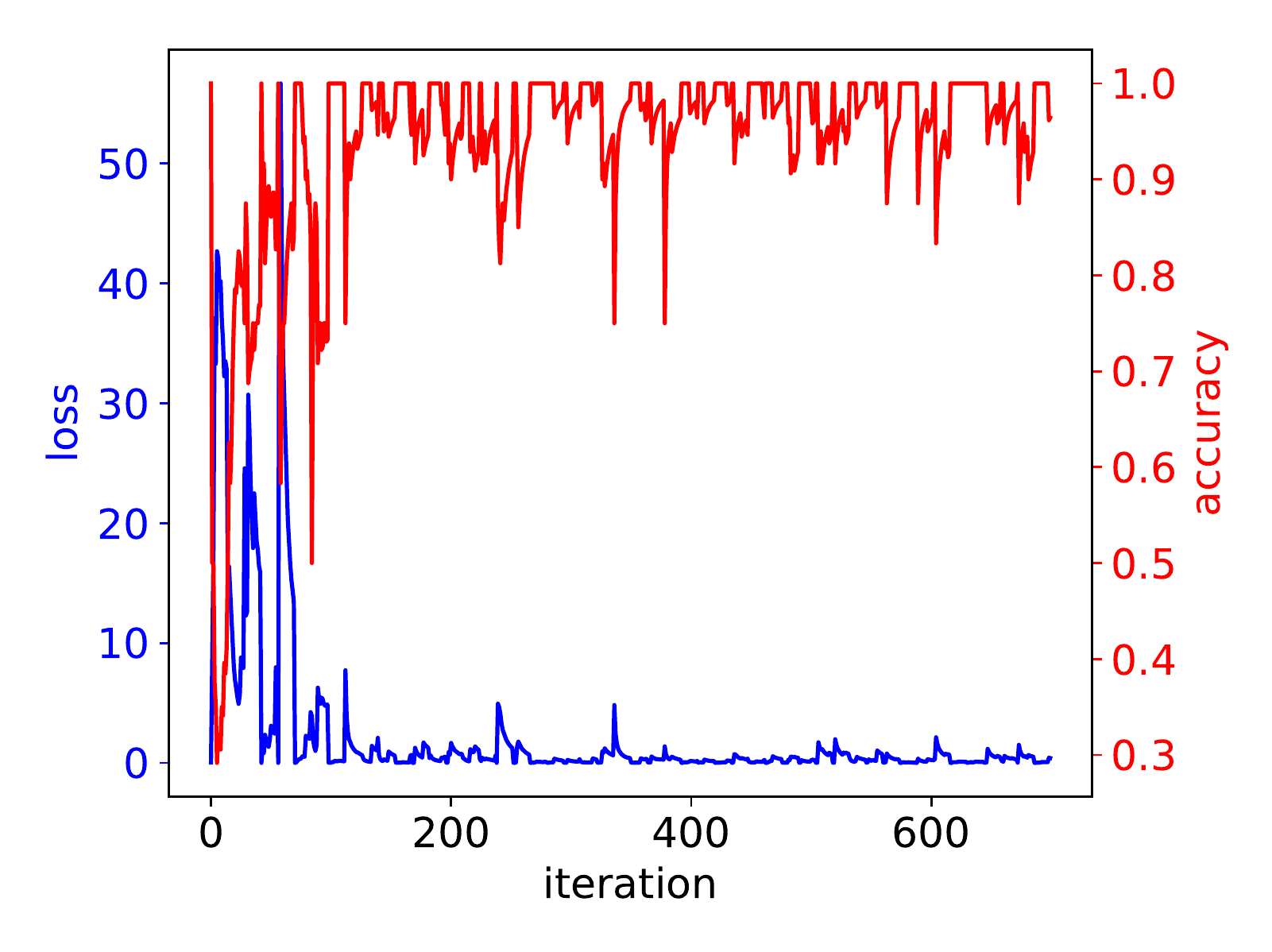}
        \caption{Max Pooling Result}
        \label{fig: maxpool}
    \end{subfigure}
    \caption{2D/3D Image Capturing}
    \label{fig: train}
    \vspace{-0.25in}
\end{figure}

\hspace{-0.15in} \textbf{Train DNN:} The DNN is trained with mini-batch strategy to make it converge more smoothly. We use $CrossEntropyLoss$ as loss function shown in equation \ref{equ: cross}. Where $x$ is output of DNN, whose dimension is $(minibatch size,2)$, and $label$ is labels for one minibatch data with dimension $(minibatch size,1)$. We use SGD optimizer to update parameters with $learning\ rate=0.01$ and $momentum=0.9$, and a $lr\ schedular$ is applied to adjust learning rate with $step size=7$ and $gamma=0.1$. Then we compare running loss and accuracy of each iterations in figure \ref{fig: train}.  Note that running loss and accuracy will be cleared after one epoch.
\begin{equation}
loss(x,label) = -log(\frac{e^{x[label]}}{\sum_je^{x[j]}})
\label{equ: cross}
\end{equation}
Figure \ref{fig: avgpool} shows the original performance of resnet18 and figure \ref{fig: maxpool} is our DNN result. It is clearly to see our DNN converges faster and has less strong vibration compare to original resnet18.

\section{Performance}\label{sec : performance}
The whole process of \textit{HRCIF} results in figure \ref{fig: final}. The first row records real human motions, the second row is the results before filtering, and the third row is a final result of detecting human. At the very beginning, human is standing on the right of radar with a wide angle of $60\degree$, where the cube in row 2 and 3 stand around $\phi=60\degree$ and $R= 110cm$. With human moves close to radar from frame $1-3$, our captured images show wide angle $\phi$ and direct distance $R$ are decreasing gradually.   While human move away from radar, the wide angle $\phi$ and direct distance $R$ are increasing. During this time, the frame ahead of last frame is "bad frame", so our DNN detects and recognizes the "misunderstanding", thus hold previous frame to the current one.
More experiments are designed to see if captured stream can be used to recognize human activities such as walk, run, jump and fall. Our results show all activities stream captured by \textit{HICFR} can be easily recognized by human with accuracy more than 90\%.

\section{Conclusion}
In general, we propose a real-time 3D human image capturing scheme based on radar, this solution not only localizes human position precisely in stream, but also protect human's privacy. Different human activities captured by our radar system  results can be easily recognized by human eyes. Our future work will focus on designing a Recurrent Neural Network (RNN) to recognize human activities with our visualization result in real-time.

\bibliographystyle{unsrt}
\bibliography{def.bib}
\end{document}